\documentclass[letterpaper, 10 pt, conference]{ieeeconf}  
\IEEEoverridecommandlockouts                             
\usepackage{graphicx}
\usepackage{amsmath,amsfonts} 
\usepackage{cite}  
\usepackage{url}

\usepackage{colortbl}
\usepackage{multirow}
\usepackage{booktabs} 
\usepackage{tabularx}
\usepackage{pbox}
\usepackage[svgnames,table]{xcolor}

\usepackage{flushend}
\usepackage{soul}
\usepackage{picinpar}
\usepackage{alltt}
\usepackage{enumerate}
\usepackage{siunitx}
\usepackage{epstopdf}
\usepackage{placeins} 
\usepackage{float}    
\usepackage[absolute,overlay]{textpos}
\usepackage{pifont}
\usepackage{bm}
\usepackage{booktabs}
\usepackage{makecell}

\usepackage{algorithmic}
\usepackage{algorithm}

\usepackage{amsthm}

\newtheorem{definition}{Definition}
\newtheorem{problem}{Problem}
\theoremstyle{remark}

\title{\LARGE \bf
Rigidity-Based Multi-Finger Coordination for Precise In-Hand Manipulation of Force-Sensitive Objects
}

\author{Xinan Rong, Changhuang Wan, Aochen He, Xiaolong Li and Gangshan Jing
\thanks{This work was supported in part by the National Key Research and Development Program of China under Grant 2025YFA1018800, the National Natural Science Foundation of China under grant (62572068, 62533002). (Corresponding author: Gangshan Jing)} 
\thanks{Xinan Rong, Aochen He, Xiaolong Li, and Gangshan Jing are with School of Automation, Chongqing University (e-mail: rongxinan2023@163.com; riemhac@163.com; 20222032@stu.cqu.edu.cn; jinggangshan@cqu.edu.cn).} 
\thanks{Changhuang Wan is with Department of Mechanical Engineering, City University of Hong Kong (e-mail: changwan@cityu.edu.hk).}
}

\allowdisplaybreaks

\begin{document}

\setlength{\textfloatsep}{10pt}
\maketitle
\thispagestyle{empty}
\pagestyle{empty}
\begin{abstract}
Precise in-hand manipulation of force-sensitive objects typically requires judicious coordinated force planning as well as accurate contact force feedback and control. Unlike multi-arm platforms with gripper end effectors, multi-fingered hands rely solely on fingertip point contacts and are not able to apply pull forces, therefore poses a more challenging problem. Furthermore, calibrated torque sensors are lacking in most commercial dexterous hands, adding to the difficulty. To address these challenges, we propose a dual-layer framework for multi-finger coordination, enabling high-precision manipulation of force-sensitive objects through joint control without tactile feedback. This approach solves coordinated contact force planning by incorporating graph rigidity and force closure constraints. By employing a force-to-position mapping, the planned force trajectory is converted to a joint trajectory. We validate the framework on a custom dexterous hand, demonstrating the capability to manipulate fragile objects—including a soft yarn, a plastic cup, and a raw egg—with high precision and safety.
\end{abstract}

\section{INTRODUCTION}
In-hand manipulation with multi-finger robotic hands has become a pivotal area in robotics research, with the aim of replicating the dexterity and precision of human hands in complex tasks\cite{billard2019trends}. Relevant application areas include medical procedures, industrial assembly, and assistive technologies for the disabled\cite{taylor2016medical, kyrarini2019robot, cook2005school}. Despite significant progress, robotic hands still suffer from low manipulation precision due to the high complexity of multi-finger coordination, and the lack of sufficient perceptual capabilities. 

To deal with the multi-finger coordination problem, optimization modeling can be employed to integrate kinematic constraints, trajectory planning, and collision avoidance \cite{sundaralingam2019relaxed, yu2025robotic}. Other approaches, such as integrating mechanism analysis with particle filtering, enhance self-recognition and control capabilities \cite{hang2021manipulation, grace2024direct}. Data-driven methods like reinforcement learning \cite{andrychowicz2020learning, suresh2024neuralfeels} and imitation learning \cite{wei2024learning} further enable complex dexterous manipulation through policy optimization. However, these works focus only on hand joints control, leaving a significant gap in addressing the precise manipulation of force-sensitive objects, which require meticulous force planning.

Force planning in manipulation traditionally relies on modeling contact constraints \cite{buss1996dextrous} and applying force optimization to evaluate feasibility \cite{prattichizzo2013motion}. Learning-based approaches, by contrast, embed force optimization into policy learning \cite{hu2025dexterous}. In addition, tactile sensors provide crucial feedback for force control and have gained extensive attention \cite{yan2021soft, lepora2022digitac}. However, their high cost, complex integration, signal noise, and processing delays limit wide practical deployment. Furthermore, most of the commercial dexterous hands lack torque control \cite{sharma2014shadow, shaw2023leap}, which further complicates precise force management.

\begin{figure}[t!]
    \centering
    \includegraphics[height=6cm]{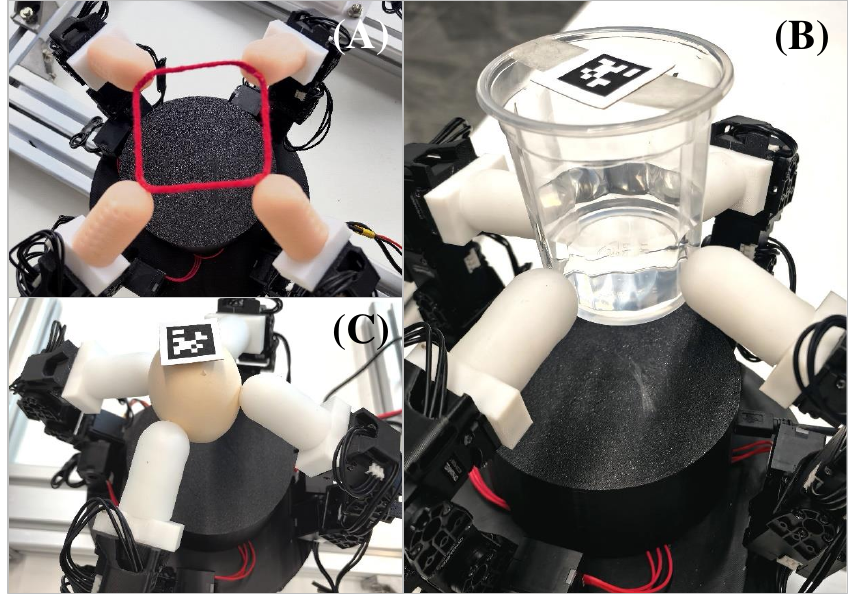}
    \caption{\small Precise in-hand movement of force-sensitive objects: (A) Manipulating a soft yarn while maintaining its shape during the movement process. (B) Manipulating a disposable plastic cup to move the AprilTag along a predefined trajectory. (C) Manipulating a raw egg to move the AprilTag along a predefined trajectory. }  
    \label{fig 3 examples}
    \vspace{-0.3cm} 
\end{figure}

Under the condition that certain physical properties of the object are known, this letter proposes a novel framework for precise in-hand trajectory tracking of force-sensitive objects without relying on tactile sensors or torque control. Three typical examples are shown in Fig. \ref{fig 3 examples} to demonstrate the problem of interest. Our approach integrates visual feedback with dynamics models to estimate and regulate applied forces. By leveraging computer vision to track the real-time pose and movement of the object, combined with model-based control strategies, robotic hands dynamically adjust applied contact forces to ensure precise trajectory tracking. We validate our method through physical experiments on a custom-designed robotic hand, demonstrating significant improvements in manipulation accuracy and reduced risk of object damage compared to conventional approaches. 

The main contribution of this letter are as follows.

(i). We propose a novel graph-rigidity-based approach for multi-finger force planning that maintains stable and safe grasp for force-sensitive objects during dynamic manipulation and remains scalable as the finger count increases.

(ii). We achieve precise force control and in-hand trajectory tracking based on artificially designed penetration distances without using torque control or tactile sensors.

\section{RELATED WORK}
\textit{1) Precise In-hand Object Movement:} For underactuated robotic hands, studies such as \cite{hang2020hand, morgan2020object, hang2021manipulation, morgan2021vision, chanrungmaneekul2023non} leverage their passive compliance to address manipulation problems under low-perception conditions. In contrast, research on fully actuated robotic hands \cite{cruciani2020benchmarking, sundaralingam2017relaxed, sundaralingam2019relaxed} has proposed kinematics-based approaches, employing optimization techniques to achieve precise control. More recent work \cite{yu2025robotic} further enhances these optimization algorithms and introduces a strategy tailored for operations involving large-area rolling contact. However, these methods primarily focus on force-tolerant objects and depend on a firm initial grasp to ensure stability. This lack of dynamic contact force adjustment makes it particularly challenging to handle force-sensitive objects effectively.

\textit{2) Coordination of End-Effectors:} Distributed formation control has been studied for multi-manipulator systems \cite{wu2022distributed, peng2023distributed}, and internal force regulation for grippers has been addressed in \cite{verginis2022cooperative}. Contact mechanics of compliant fingertips were analyzed in \cite{smith2012dual}, and distributed transportation via rolling contacts was shown in \cite{cortez2023distributed}. In contrast, multi-finger collaboration involves small contact patches, limited friction margins, and a highly coupled internal force space, leading to stricter requirements on contact stability and force coordination for force-sensitive objects.

\textit{3) Learning-based Manipulation:} 
Learning-based methods have demonstrated remarkable capabilities in complex in-hand manipulation tasks, achieving implicit force control \cite{andrychowicz2020learning, suresh2024neuralfeels, wei2024learning, hu2025dexterous}, while recent work \cite{zeng2022multifingered} has mapped visual inputs to force control through deep learning. However, the force control processes in these approaches generally lack interpretability and fail to elucidate the underlying physical mechanisms. Some studies simplify tactile signals to binary contact representations \cite{lee2024dextouch, yuan2024robot}, while \cite{yang2024anyrotate} extends tactile representation to continuous physical quantities. Nevertheless, these methods typically rely on complex sensor calibration, are primarily designed for rigid objects, and still lack interpretability in force modeling.

\section{Dynamics Modeling and Problem Statement}

\begin{figure}[t]  
    \centering
    \includegraphics[height=5cm]{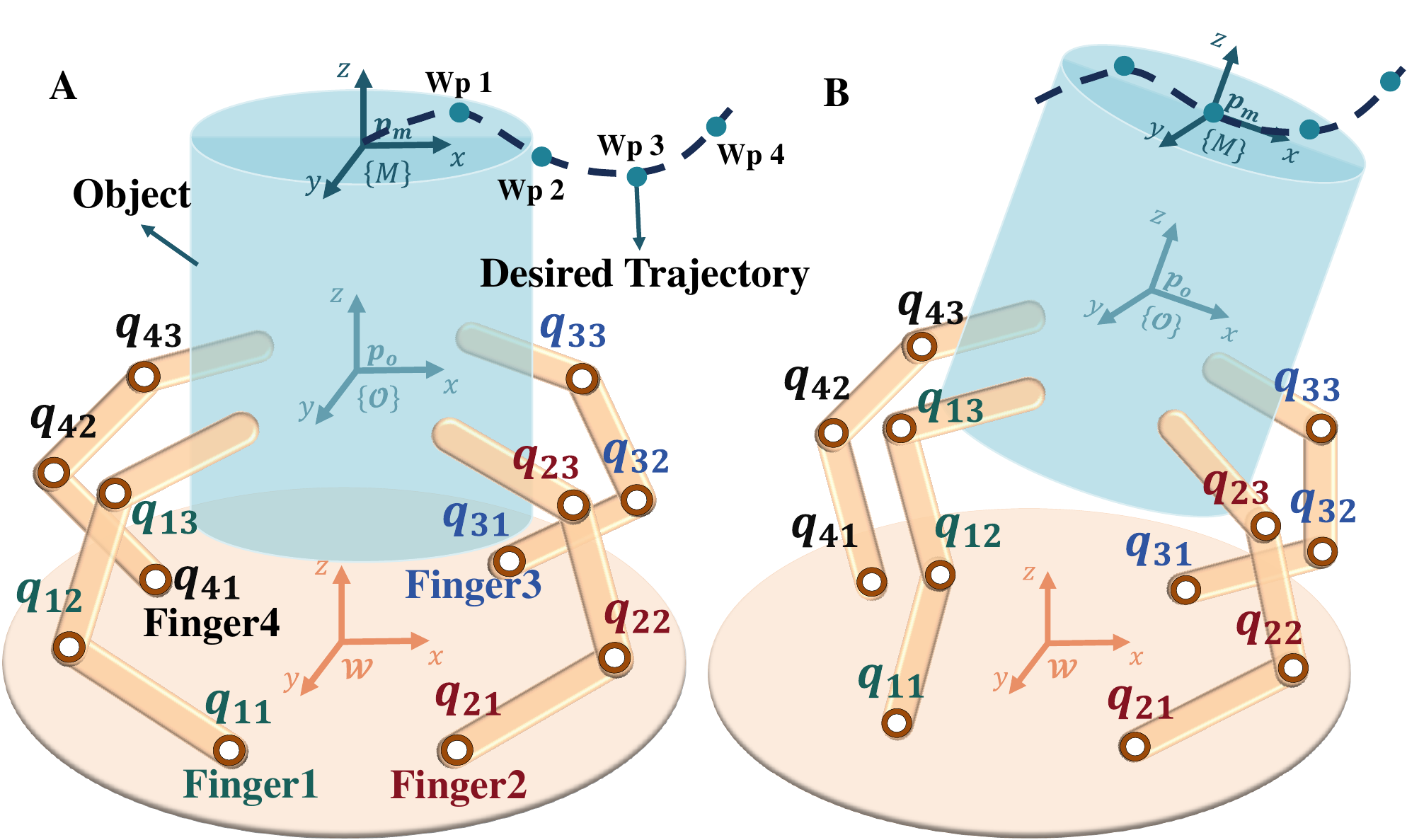}  
    \caption{\small Precision in-hand object movement: (A) The object moves along a predefined trajectory, with coordinate system $\{\mathcal{M}\}$ following each waypoint ($Wp1, \dots$), achieving precise motion through joint position control. (B) The motion process of manipulation frame $\{\mathcal{M}\}$.}  
    \label{fig:motion_frames}
    \vspace{-0.3cm} 
\end{figure}

We consider a dexterous manipulation system with $m \geq 3$ fingers (indexed by $i \in M$) and $n$ joints per finger. As shown in Fig.~\ref{fig:motion_frames}, we define four coordinate frames: the \textbf{Global frame} $\{\mathcal{W}\}$; the \textbf{Object frame} $\{\mathcal{O}\}$ at the center of mass (CM) $\mathbf{p}_o$; the \textbf{Manipulation frame} $\{\mathcal{M}\}$ at $\mathbf{p}_m$ for visual tracking; and the \textbf{Contact frames} $\{\mathcal{C}_i\}$ at fingertips $\mathbf{p}_i$.

\subsection{Hand-Object System Modeling}
The object dynamics is given as follows \cite{pfanne2022hand}:
\begin{equation}
	\mathcal{W}_{inertial} + g_o = \mathcal{W}_o = G f_c \label{Dynamics of the object},
\end{equation}
where $\mathcal{W}_{inertial} = M_o \dot{\mathcal{V}}_o + C_o \mathcal{V}_o$ is the inertial wrench with $\mathcal{V}_o \in \mathbb{R}^6$ the object twist, $g_o, \mathcal{W}_o \in \mathbb{R}^6$ are the gravitational and total external wrenches, respectively.
The grasp matrix $G \in \mathbb{R}^{6 \times 3m}$ maps contact forces $f_c \in \mathbb{R}^{3m}$ to the total wrench, constructed from blocks $G_i = [I_3, S(\mathbf{p}_{io})]^\top$, where $\mathbf{p}_{io} = \mathbf{p}_i - \mathbf{p}_o$ is the vector from object center to \textit{i}-th contact point, and $S(\cdot)$ is the skew-symmetric matrix operator. 
Considering the manipulation of lightweight objects, the object inertia matrix $M_o$ and Coriolis terms $C_o$ are small, rendering the inertial wrench negligible ($\mathcal{W}_{inertial} \approx 0$). This approximation eliminates the need for precise inertial parameters. Planning desired contact forces $f_c$ for gentle, stable manipulation is the key challenge.

Since most commercial dexterous hands lack direct torque control, our approach operates in the joint position domain. Nevertheless, to characterize the dynamic coupling between contact forces and fingertip motions, the task-space inertia matrix $M_c$ is required. Combining the differential kinematics $v_c = J_h\dot{q}$ with the standard joint-space dynamics yields:
\begin{equation}
	M_c = (J_h^+)^\top M(q) J_h^+ \label{task_inertia},
\end{equation}
where $v_c \in \mathbb{R}^{3m}$ and $\dot{q} \in \mathbb{R}^{mn}$ are the contact point and joint velocities, $M(q) \in \mathbb{R}^{mn \times mn}$ is the joint-space inertia matrix, $J_h(q) \in \mathbb{R}^{3m \times mn}$ is the hand Jacobian, and $(\cdot)^+$ denotes the Moore--Penrose pseudoinverse.

\subsection{Problem Statement}
Given the dynamics of an $m$-fingered robotic hand and its interaction with a grasped object, the goal is to control a force-sensitive, elastically deformable object so that the manipulation frame $\{\mathcal{M}\}$ follows a desired trajectory while maintaining the optimal initial contact configuration, as measured by the grasp quality metric~\cite{roa2015grasp}.

Formally, the problem can be stated as follows.
\begin{problem}
Given a desired trajectory \(x_d(t)\) of the object pose in \(t \in [0, T]\), 
design a joint trajectory \(q(t)\) such that:  
(i) the actual object trajectory \(x(t)\) tracks \(x_d(t)\), with the pose tracking error 
\(\|x(t) - x_d(t)\| \leq \delta\) for all \(t \in [0, T]\), where \(\delta > 0\) is a threshold for the allowed transient pose error; 
(ii) the contact forces $f_c$ ensure a stable grasp and satisfy the safety constraints that prevent object deformation.
\end{problem}

\subsection{Distinctions from Multi-Arm Coordination}

While multi-finger manipulation shares coordination principles with multi-arm systems, it differs fundamentally in contact mechanics and force constraints.

First, unlike multi-arm models assuming torque-transmitting rigid contacts \cite{smith2012dual}, we model fingertips as frictional point contacts strictly bounded by Coulomb constraints. Consequently, our grasp matrix $G$ is formulated for point contacts, differing from the Jacobian-based formulations in rigid-link systems (e.g.,\cite{smith2012dual}, Eq.~(1)).

Second, the role of internal forces is reversed. In contrast to multi-arm manipulation strategies that suppress internal forces (e.g., the “no-internal-force” approach in \cite{verginis2022cooperative}), multi-finger manipulation requires to design nonzero and tightly coupled internal forces for stable grasp. As the number of fingers increases, the interdependence of these forces grows, limiting the scalability of force-analysis methods designed for dual-arm systems (such as \cite{smith2012dual}). We address this by introducing a rigid-contact framework that systematically characterizes constraints while remaining scalable with respect to finger count.

Finally, research on force-sensitive objects remains sparse in multi-arm coordination. This work fills this gap by achieving precise internal-force regulation on fragile objects.

\section{Method}
\begin{figure}[t]  
    \vspace{-0.3cm}
    \centering
    \includegraphics[height=4.8cm]{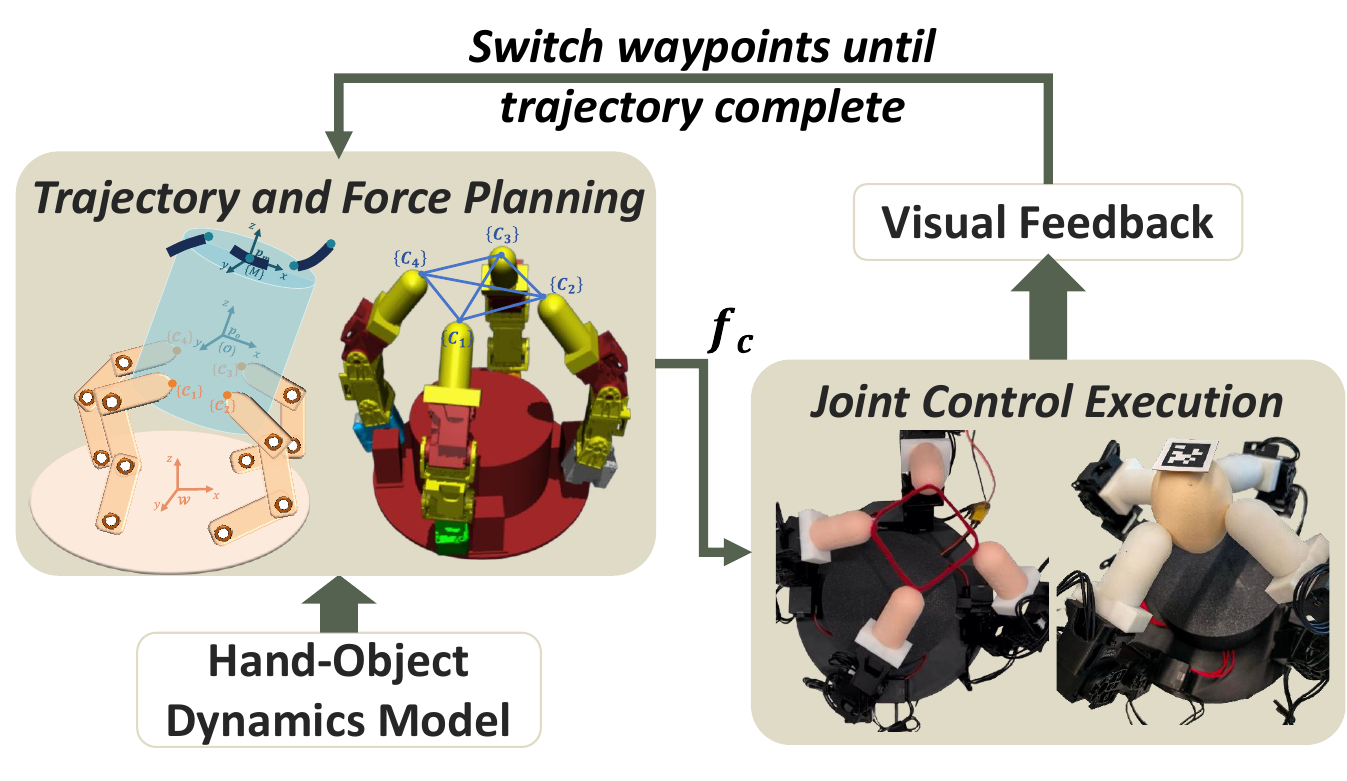}  
    \caption{RFP-FJC Algorithm Framework}  
    \label{RFP-FJC Algorithm Framework}
    \vspace{-0.3cm}
\end{figure}

To address the challenges of stable in-hand manipulation for force-sensitive objects, we propose a novel dual-layer algorithm consisting of a rigidity-based force planner and a force-to-joint converter (RFP-FJC), 
as illustrated in Fig.~\ref{RFP-FJC Algorithm Framework}.

Next, we will introduce the details of the two layers in Subsections \ref{subsec rigidity} - \ref{subsec force optimization} and Subsection \ref{subsec excecution}, respectively.

\subsection{Rigid Contact Framework}\label{subsec rigidity}
The primary challenge in manipulating force-sensitive objects is coordinating coupled internal forces for stability without causing deformation. Our \textit{Rigid Contact Framework} addresses this by recasting the complex force problem into a more tractable kinematic one. It imposes a geometric constraint in $\mathbb{R}^3$—the invariance of Euclidean distances between contact points, implicitly regulating these internal forces.

\begin{definition}
The contact framework of an \(m\)-fingered robotic hand  in \( \mathbb{R}^3 \) is defined as \( F = (\mathcal{G}, p_\mathcal{G}) \),  where \( \mathcal{G} = (V, E) \) is a  graph with vertices \( V =\{1,...,m\}\) representing contact points; each edge $(i, j) \in E$ encodes a distance constraint between contact points $i$ and $j$; the configuration $p_{\mathcal{G}}= [p_1^\top, p_2^\top, \dots, p_m^\top]^\top: V \rightarrow \mathbb{R}^{3m}$ corresponds to the spatial positions of the $m$ contact points.
\end{definition}

This rigidity constraint is expressed mathematically as: 
\[  \|p_i(t) - p_j(t)\| = \|p_i(0) - p_j(0)\|, \quad \forall (i,j) \in E,~t\geq0. \] 

\begin{figure}[htbp]  
    \vspace{-0.3cm}
    \centering
    \includegraphics[height=3.7cm]{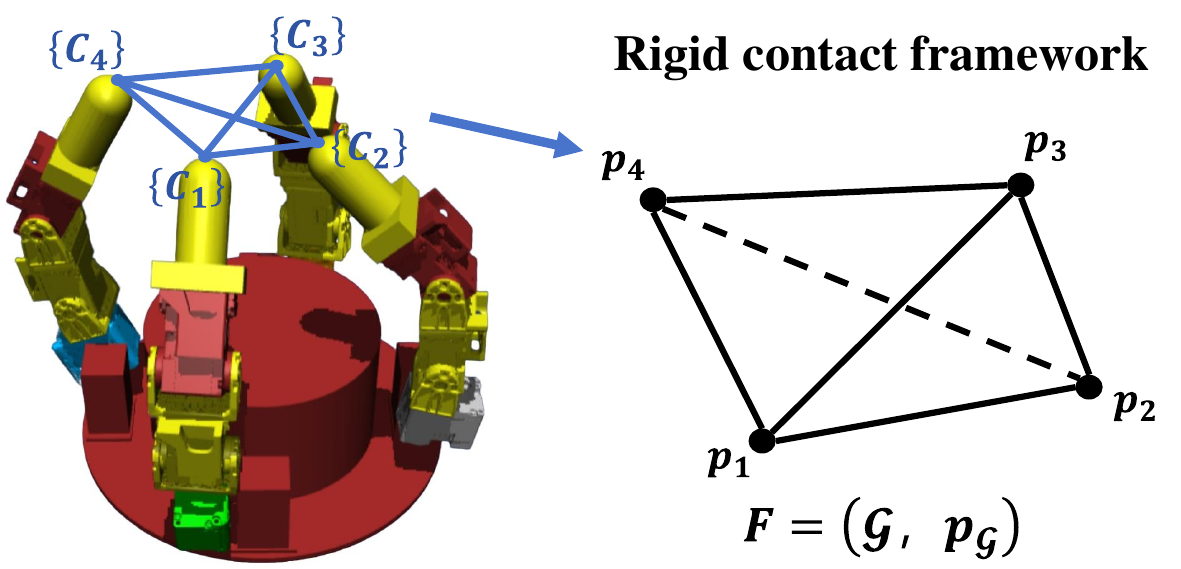}  
    \caption{\small A graph-based abstraction of geometric constraints among contact points, modeling a rigid configuration that remains invariant during grasp execution.}  
    \label{fig:rigid_framework}
    \vspace{-0.3cm}
\end{figure}

Fig.~\ref{fig:rigid_framework} shows the case with four contact points exemplifying a rigid contact framework in three-dimensional space. To guarantee rigidity of the framework, \( \mathcal{G} = (V, E) \) should be a complete undirected graph \cite{hendrickson1992conditions}. For cases with more contact points, however, it is not always necessary for $\mathcal{G}$ to be a complete graph in order to maintain rigidity. 

The rigidity function of the contact framework is defined based on pairwise distances between contact points:
\[
\phi = [..., \| p_i - p_j \|^2, ...]^\top,\quad \forall (i, j) \in E.
\]

Given a contact framework $F$, the infinitesimal motions of the contact points are defined as $v$ such that $p_\mathcal{G}$ preserves $\phi$ if $\dot{p}_\mathcal{G}=v$. That is, 
\begin{equation}
    \dot{\phi}=\frac{\partial \phi}{\partial p_\mathcal{G}}v=0.
\end{equation}

To ensure a stable grasp, all the infinitesimal motions of contact points are desired to be trivial motions, which correspond to the six rigid-body motions (uniform translations and rotations of the whole framework). Therefore, one has
\begin{equation}\label{rank R}
    {\rm rank}(R) = 3m - 6,
\end{equation}
where $R =\frac{\partial \phi}{\partial p_\mathcal{G}}$ is called the rigidity matrix. According to \cite{hendrickson1992conditions}, the rank condition \eqref{rank R} holds as long as the contact framework is infinitesimally rigid.

Consequently, to ensure precise tracking of the desired object trajectory, the fingertip velocity \( v_c(t) \) must satisfy
\begin{equation}
	R(x(t)) v_c(t) = 0. \label{rigid constraint}
\end{equation}

Given that the contact point position $p_{i,\mathcal{O}}$ in the object frame $\{O\}$ remains constant after the initial grasp, the object's pose trajectory $x(t)$ yields the homogeneous transformation matrix ${}^{\mathcal{W}}\mathbf{T}_{\mathcal{O}}(t)$ from $\{\mathcal{O}\}$ to the world frame $\{\mathcal{W}\}$. Consequently, the position of the contact point in the world frame is given by $[p_{i,\mathcal{W}}^\top,1]^\top={}^{\mathcal{W}}\mathbf{T}_{\mathcal{O}}(t)[p_{i,\mathcal{O}}^\top,1]^\top$, thus converting the trajectory of the object $x(t)$ into the trajectory of the contact point $p(t)$ in the world frame.

\subsection{Contact Force Constraints and Decomposition}
\label{subsec force constraints}

To ensure stable manipulation without causing damage, the contact force $f_c = [f_{c_1}^\top, \ldots, f_{c_m}^\top]^\top$ must satisfy a set of critical physical constraints.

\begin{figure}[t]  
    \centering
    \includegraphics[height=5.8cm]{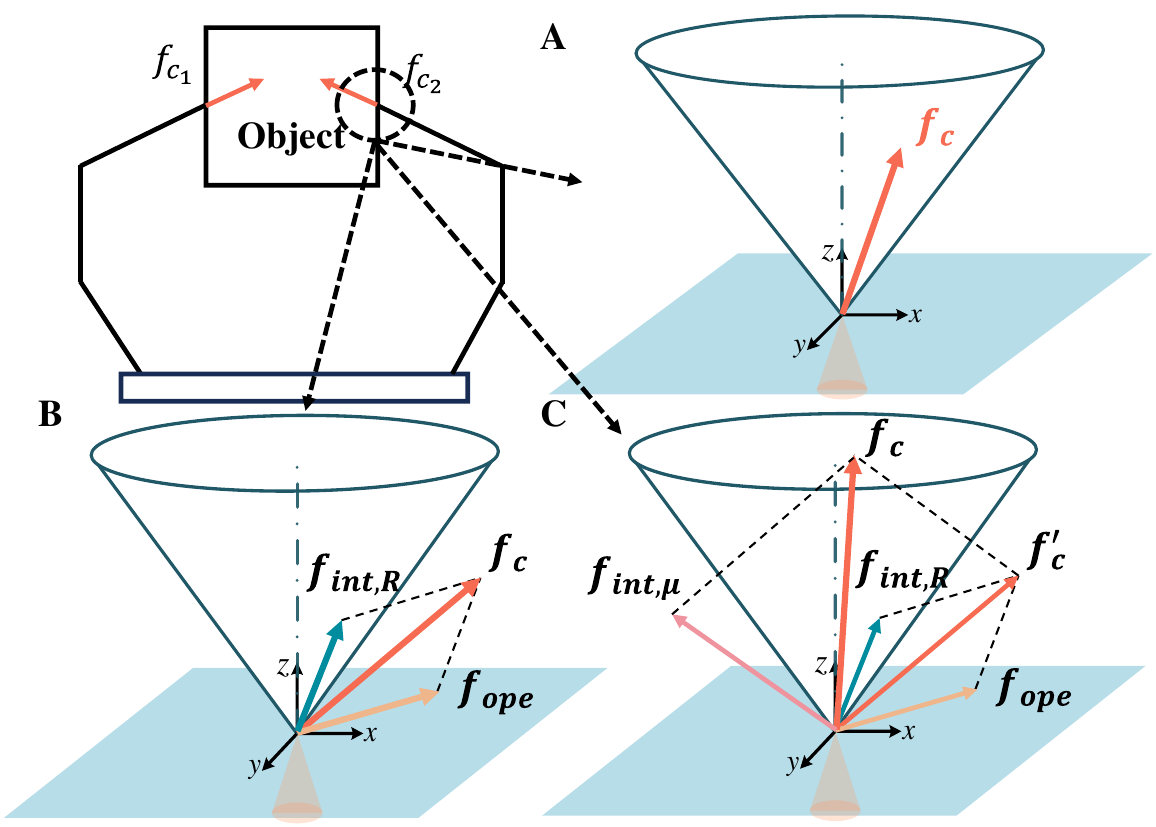}  
    \caption{\small Contact force analysis: (A) Contact force \( f_c \)  lies within the friction cone, indicating stability. (B) With only \( f_{\text{ope}} \) and \( f_{\text{int},R} \), the resultant falls outside the cone. (C) The combined force of \( f_{\text{ope}} \), \( f_{
    \text{int},R} \), and \( f_{\text{int},\mu} \) lies well within the cone, ensuring a stable grasp.}  
    \label{fig:force_analysis}
    \vspace{-0.3cm}
\end{figure}

First, to prevent slippage, the contact force $f_{c_i}$ at each fingertip must lie within its corresponding friction cone (Fig.~\ref{fig:force_analysis}A). This imposes the following constraint on its components in the local contact frame $\{\mathcal{C}_i\}$:
\begin{equation} \label{Force closure condition}
	\sqrt{f_{c_{ix}}^2 + f_{c_{iy}}^2} \leq \mu f_{c_{iz}}
\end{equation}
where $f_{c_i} = [f_{c_{ix}}, f_{c_{iy}}, f_{c_{iz}}]^\top$ and $\mu$ is the static friction coefficient.

Second, to prevent object damage while ensuring a stable grip, the normal component of the contact force, $f_{c_{iz}}$, must be maintained within a predefined safe range:
\begin{equation} \label{Force range}
	f_{n, \min} \leq f_{c_{iz}} \leq f_{n, \max}
\end{equation}
Here, $f_{n, \min}$ is the minimum normal force required to secure the object, while $f_{n, \max}$ is the maximum allowable force to prevent its deformation.

Satisfying these constraints requires a trade-off between grasp stability and object safety. Specifically, preventing slippage (Eq.~\eqref{Force closure condition}) favors higher normal forces to maximize the friction margin, whereas preventing deformation (Eq.~\eqref{Force range}) strictly limits their magnitude. Directly computing a contact force $f_c$ that balances these competing objectives is computationally challenging.

To address this, we propose a decoupled force decomposition strategy. Unlike standard approaches that treat internal forces as a unified variable~\cite{verginis2022cooperative}, we explicitly partition the contact force into three functional components to isolate geometric constraints from physical constraints:
\begin{equation} \label{Force decoupling}
	f_c = f_{\text{ope}} + f_{\text{int},R} + f_{\text{int},\mu}
\end{equation}
subject to the equilibrium constraints:
\begin{equation} \label{Null space constraints}
    f_{\text{ope}} = G^+ \mathcal{W}_{\text{ext}}, \quad G f_{\text{int},R} = \mathbf{0}, \quad G f_{\text{int},\mu} = \mathbf{0}.
\end{equation}

These components serve distinct, complementary roles:

(i) \textbf{Operational Force} ($f_{\text{ope}}$): Balances the external wrench (e.g., gravity). While necessary for equilibrium, this component is determined solely by object dynamics and may violate friction constraints.

(ii) \textbf{Rigidity Internal Force} ($f_{\text{int},R}$): Analytically derived to enforce the rigid contact framework. It ensures that the relative motion of contact points complies with the rigidity constraint (Eq.~\eqref{rigid constraint}) without disturbing the object's trajectory.

(iii) \textbf{Friction Internal Force} ($f_{\text{int},\mu}$): A supplementary component optimized to satisfy friction cones and safety limits (Eqs.~\eqref{Force closure condition}-\eqref{Force range}). It is introduced only when the combination of $f_{\text{ope}}$ and $f_{\text{int},R}$ is insufficient to guarantee stability, as visually demonstrated in Fig.~\ref{fig:force_analysis}.

Based on the decomposition \eqref{Force decoupling}, the planning of the contact force $f_c$ can be converted into tractable sub-problems.

\subsection {Contact Forces Optimization}\label{subsec force optimization}
Following the three-part decomposition, we now detail the computation for each component. 

Based on the simplified dynamics model Eq.~\eqref{Dynamics of the object} established in Section III-A, the $f_{\text{ope}}$ is given by:
\begin{equation}
	f_{\text{ope}} = - G^+ g_o. \label{f_ope_computation}
\end{equation}

To enforce the rigidity constraint at the acceleration level, we differentiate the velocity-level equation \( R v_c = 0 \). This step is justified by the explicitly satisfied initial condition \( R(x(t_0)) v_c(t_0) = 0 \), which yields:
\begin{equation}
    R \dot{v}_c + \dot{R} v_c = 0 \label{rigid constraint dot}.
\end{equation}

The rigidity internal force, \(f_{\text{int}, R}\), is determined by solving the following optimization problem:
\begin{equation}
\begin{aligned}
\label{eq.prob}
\min_{\dot{v}_c}\quad & (\dot{v}_c - \alpha)^\top M_c (\dot{v}_c - \alpha) \\
\text{s.t.}\quad & R \dot{v}_c + \dot{R} v_c = 0,
\end{aligned}
\end{equation}
where \( \alpha \) represents the acceleration of the unconstrained motion in the task space (i.e., the case where the hand does not grasp the object and thus no kinematic contact force constraints are imposed). The solution is obtained from the first-order necessary conditions for optimality, yielding:
\[
M_c \dot{v}_c = M_c \alpha - R^\top (R M_c^{-1} R^\top)^{+} (\dot{R} {v}_c + R \alpha).
\]

The internal force \(f_{\text{int}, R}\) is thus given by:
\setlength{\abovedisplayskip}{4pt} 
\setlength{\belowdisplayskip}{4pt} 
\begin{equation}
    f_{\text{int}, R} = M_c (\alpha - \dot{v}_c) = R^\top (R M_c^{-1} R^\top)^{+} (\dot{R} v_c + R \alpha).\label{f_{int,R}}
\end{equation}

With the rigid component established, the frictional internal force, \(f_{\text{int},\mu}\), is found by solving an optimization problem that minimizes the force required to prevent slip. This problem is formulated with a nonlinear objective and linear constraints as follows:
\begin{subequations} \label{fric_opt}
\begin{align}
  \min_{f_{\text{int}, \mu}}~~~~ & \frac{1}{2} \left( \frac{f_{c,\parallel}}{\mu f_{c,\perp}} \right)^2 \label{eq:fric_opt_obj} \\
  \text{s.t.}~~~~     & f_{c,\perp} \geq \mathbf{0}, \label{eq:fric_opt_compressive} \\
                      & G f_{\text{int},\mu} = \mathbf{0}, \label{eq:fric_opt_nullspace} \\
                      & f_c = f_{\text{ope}} + f_{\text{int},R} + f_{\text{int},\mu}, \label{eq:fric_opt_balance} \\
                      & \mu f_{c,\perp} - f_{c,\parallel} \geq \mathbf{0}, \label{eq:fric_opt_cone} \\
                      & f_{n, \text{min}} \leq f_{c,\perp} \leq f_{n, \text{max}}, \label{eq:fric_opt_bounds}
\end{align}
\end{subequations}
where \( f_{c,\parallel} \) and \( f_{c,\perp} \) denote the tangential and normal components of the contact force \(f_c\) in the local contact frame, respectively. The objective function \eqref{eq:fric_opt_obj} maximizes the margin to the friction cone boundary by minimizing the ratio of the tangential force to the maximum allowable static friction force. Specifically, constraint \eqref{eq:fric_opt_compressive} ensures a compressive contact force; \eqref{eq:fric_opt_nullspace} restricts \( f_{\text{int},\mu} \) to the null space of the grasp matrix, ensuring it is a pure internal force; \eqref{eq:fric_opt_cone} enforces the Coulomb friction cone constraint; and \eqref{eq:fric_opt_bounds} bounds the normal component of the contact force to be within pre-defined operational limits.

The details of the rigidity-based force planner introduced in this section are summarized in Algorithm \ref{alg:force_planner}.
\begin{algorithm}[t]
	\caption{Rigidity-Based Force Planner}
	\label{alg:force_planner}
	\renewcommand{\algorithmicrequire}{\textbf{Input:}}
	\renewcommand{\algorithmicensure}{\textbf{Output:}}
	\begin{algorithmic}[1]
		\REQUIRE $x_c, \mu, f_{n, \text{min}}, f_{n, \text{max}}, g_o$ \COMMENT{\small contact config., friction coef., force limits, object dynamics}
		\ENSURE  $f_c$    
		
		\STATE $(M, J_h) \leftarrow \text{GetRobotDynamics}()$ \\
        \COMMENT{\small Obtain robot dynamics}
        \STATE $M_c \leftarrow \text{GetTaskInertia}(M, J_h)$ \COMMENT{\eqref{task_inertia}} 
		\STATE $G  \leftarrow \text{CalcGraspMatrix}(x_c)$ 
        \COMMENT{\small Grasping matrix calculation}
		\STATE $f_{\text{ope}} \leftarrow \text{CalcOpeForce}(G, g_o)$ \COMMENT{\eqref{f_ope_computation}}
		\STATE $f_{\text{int},R} \leftarrow \text{CalcRigidIntForce}(x_c, M_c)$ \COMMENT{\eqref{f_{int,R}}}
        
        \STATE $f_{c,\text{pre}} \leftarrow f_{\text{ope}} + f_{\text{int},R}$
        \COMMENT{\small Preliminary force for constraint check}
        \STATE $(f_{c,\perp}, f_{c,\parallel}) \leftarrow \text{Decompose}(f_{c, \text{pre}})$ \\
        \COMMENT{\small Get normal and tangential components}
        
        \STATE $f_{\text{int},\mu} \leftarrow \mathbf{0}$ \COMMENT{\small Initialize frictional internal force to zero}
	
		\IF{$\|f_{c,\parallel}\|>\mu f_{c,\perp}$ \text{or} $f_{c,\perp}<f_{n,\min}$ \text{or} $f_{c,\perp}>f_{n,\max}$}
		    \STATE $f_{\text{int},\mu} \leftarrow \text{CalcFricForce}(f_{\text{ope}}, f_{\text{int},R}, G, \mu, f_{n,\min}, f_{n,\max})$ \\ 
            \COMMENT{\eqref{fric_opt}}
		\ENDIF
		
		\STATE $f_{\text{int}} \leftarrow f_{\text{int},R} + f_{\text{int},\mu}$
        \COMMENT{\small Total internal force}
        \STATE $f_c \leftarrow f_{\text{ope}} + f_{\text{int}}$ \COMMENT{\small Final contact force}
        
		\RETURN $f_c$        
	\end{algorithmic}    
\end{algorithm}

\subsection{Force-to-Position Mapping and Control}\label{subsec excecution}
To realize the desired contact forces from Section \ref{subsec force optimization} without torque control or tactile sensing, we reformulate force control as a position tracking problem via a compliance-based virtual penetration model.

Specifically, for each fingertip, we define a virtual target point inside the object. Commanding the fingertip to track this point via position control generates the desired force as the object's surface resists the motion. The virtual target point for the \(i\)-th fingertip is:
\begin{equation}\label{virtual target point}
	x_{c_i,d} = x_{c_i} + D_i,
\end{equation}
where \( x_{c_i,d} \in \mathbb{R}^3 \) denotes the virtual target point, \( x_{c_i} \in \mathbb{R}^3 \) is the current contact position on the object surface, and \( D_i \in \mathbb{R}^3 \) is the penetration distance vector \cite{tang2024robotic}, which encodes the desired contact force.

The mapping between the penetration distance and the contact force is modeled as a linear relationship:
\begin{equation}\label{penetration distance}
	D_i = \text{diag}(c_x, c_y, c_z) f_{c_i},
\end{equation}
where \( c_x, c_y, c_z \) are the virtual compliance gains along the \( x, y, z \) axes. These parameters are empirically tuned based on the target object's physical properties (e.g., stiffness and surface material): smaller \( c \) values are assigned to deformable or fragile objects to prevent excessive shape distortion, while larger \( c \) values are employed for heavier or stiffer objects to ensure sufficient penetration depth for stable holding.

We define \( q_t \in \mathbb{R}^{mn} \) as the joint angles at time \( t \in [1, T] \), bounded by \( q_{\min} \) and \( q_{\max} \). Let \( (p_o, R_o) \) and \( (p_{o,d}, R_{o,d}) \) denote the actual and desired terminal object poses. For each fingertip \( i \in M \), \( (p_{fin,i,t}, R_{fin,i,t}) \) is its actual pose, and \( (x_{c_i,d,t}, R_{d,i}) \) is the desired pose, where \( x_{c_i,d,t} \) is the virtual target point derived from \( f_{c_i} \) via \eqref{virtual target point}-\eqref{penetration distance}. The joint planning problem is formulated as the following nonlinear and non-convex constrained optimization, which benefits from low dimensionality for efficient solution:
\begin{subequations} \label{Trajectory Optimization}
\begin{align}
    \min_{p_o, R_o, q_{1:T}} \quad & \lambda_1 \left( \left\| p_o - p_{o,d} \right\| + \left\| \log(R_o^\top R_{o,d}^{-1}) \right\| \right) \nonumber \\
    & + \lambda_2 \sum_{i \in M} \sum_{t=1}^{T} \left\| \log(R_{fin,i,t}^\top R_{d,i}^{-1}) \right\| 
    \label{objective} \\
    \text{s.t.} \quad & (p_{fin,i,t}, R_{fin,i,t}) = \text{FK}_i(q_t), \quad \forall t, \label{fk_kinematics} \\
    & \left\| p_{fin,i,t} - x_{c_i,d,t} \right\| \leq \epsilon, \quad \forall  t, \label{slippage_constraint} \\
    & q_{\min} \leq q_t \leq q_{\max}, \quad \forall t, \label{joint_limits}
\end{align}
\end{subequations}
where \(\lambda_1\) and \(\lambda_2\) are weighting factors. The first term penalizes the terminal pose error of the object, with \(\log(\cdot)\) mapping rotation matrices to \(\mathbb{R}^3\). The second term penalizes orientation deviations of fingertips from desired contact configurations. Constraint \eqref{fk_kinematics} computes fingertip poses via forward kinematics $\text{FK}_i(q_t)$; \eqref{slippage_constraint} ensures each fingertip remains within \(\epsilon\) of its virtual target to prevent slippage while allowing force regulation; \eqref{joint_limits} enforces joint limits.

The complete pipeline is summarized in Algorithm \ref{force-to-joint converter}, which integrates the force planner (Algorithm \ref{alg:force_planner}) with trajectory optimization \eqref{Trajectory Optimization} in an MPC \cite{garcia1989model} loop. Each iteration computes contact forces, maps them to virtual targets, solves for joint trajectories, and executes through visual feedback.

\begin{algorithm}[htbp]
	\caption{Force-to-Joint Converter}
	\label{force-to-joint converter}
	\renewcommand{\algorithmicrequire}{\textbf{Input:}}
	\renewcommand{\algorithmicensure}{\textbf{Output:}}
	\begin{algorithmic}[1]
		\REQUIRE $iter$, $x_c$, $\lambda_1$, $\lambda_2$, $\epsilon$, $q_{\min}$, $q_{\max}$, $\delta$, $x_{o,d} = (p_{o,d}, R_{o,d})$
        \COMMENT{\small max iterations, contact config., weights, tolerance, joint limits, convergence threshold, target object pose}
	
            \ENSURE  $q(1:T)$	\COMMENT{\small Joint positions}	
		
		\STATE \(\text{InitGrasp}(x_c)\)	\COMMENT{\small Form the initial grasp}	
        \STATE \(\text{LoadParams}(\mu, f_{n,\min}, f_{n,\max}, g_o, k)\) \\ \COMMENT{\small Load system parameters}
		\FOR{ $i \in [0,iter]$}		
		\STATE \(f_c \leftarrow \text{ForcePlanning}(x_c, \mu, f_{n,\text{min}}, f_{n,\text{max}}, g_o)\) \\ 
        \COMMENT{\small Algorithm \ref{alg:force_planner}}
        \STATE \(D \leftarrow \text{CalcPenDist}(k, f_c)\) \COMMENT{\eqref{penetration distance}}
		\STATE \(x_{c,d}\leftarrow \text{CalcVirtualTarget}(x_c,D)\)  \COMMENT{\eqref{virtual target point}}
		\STATE \(q(1:T)\leftarrow \text{SolveIK}(x_{c,d}, x_{o,d}) \) \COMMENT{\eqref{Trajectory Optimization}}
		\STATE \(\text{Hand.Execute}(q(1))\)\COMMENT{\small Execute first step (MPC)}
		\STATE \(x_{o,current} \leftarrow \text{VisualFeedback}()\) \COMMENT{\small Observe object pose}
		\IF{$\|x_{o,current} - x_{o,d}\| \leq \delta$} 
            \STATE \textbf{break}
        \ENDIF \COMMENT{\small Check if target pose is reached}
		\ENDFOR	
	\end{algorithmic}
\end{algorithm}

\section{Experiment}

This section presents practical experimental results validating the proposed algorithms. A video demonstration is available at \url{https://youtu.be/kcf9dVW0Dpo}.

\subsection{Hardware Setup}
The platform is a fully actuated four-finger hand modified from the LEAP Hand \cite{shaw2023leap}. 
We redesigned the arrangement by applying the original thumb structure to all fingers, ensuring consistent kinematics and independent actuation. 
To rigorously evaluate the strategy under low sensor-dependence, tactile sensors are explicitly omitted; the system relies solely on joint encoders and an Intel RealSense D405 camera (tracking AprilTags) for state estimation. 

\begin{figure}[t]  
    \centering
    \includegraphics[height=5cm]{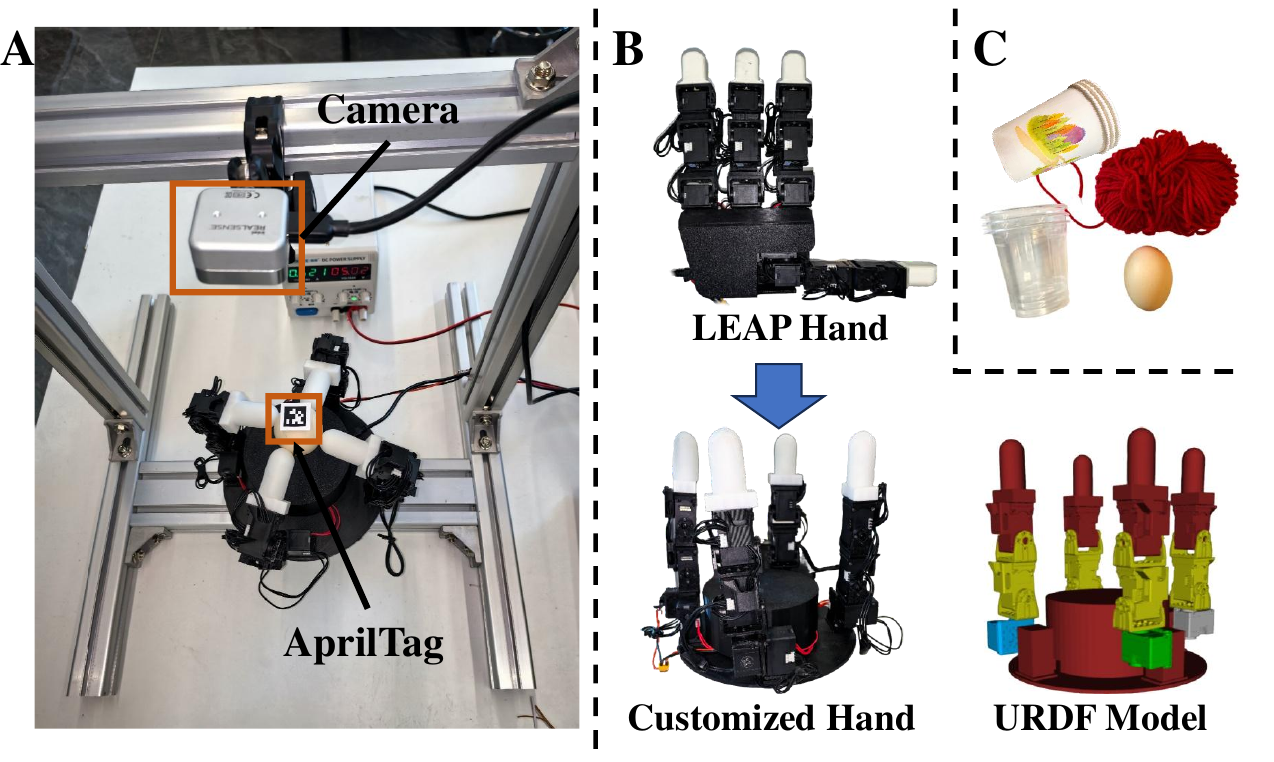}  
    \caption{\small Experimental setup: (A) System with camera positioned 35 cm above objects. AprilTag enables tracking. (B) Physical and URDF models of the customized hand. (C) Manipulated objects: disposable cup, raw egg, and yarn.}
    \label{fig:exp_setup}
    \vspace{-0.3cm}
\end{figure}

The hardware platform is shown in Fig.~\ref{fig:exp_setup}. An Intel RealSense D405 camera above the structure provides visual input, and AprilTag markers on visible surfaces enable accurate pose estimation. The system runs on Ubuntu 20.04.
For optimization, the Sequential Least Squares Programming (SLSQP) algorithm is employed. Following~\cite{yu2025robotic}, we analytically derive and directly provide the gradients of the objective and constraints, which improves computational efficiency. Experimental results show that the proposed optimization is suitable for online control. The object mass $M$ was directly measured, and the friction coefficient $\mu$ was specified according to typical material properties.

\subsection{Experiment Design and Results}

\subsubsection{Yarn-Frame Stability: Validating the Rigid Contact Model}
To validate the rigid contact model, we command a square yarn-frame formed by the fingertips to trace a square trajectory, quantifying its stability with the \textit{Total Relative Deviation} (TRD): the percentage change in inter-fingertip distances from their initial state:
\begin{equation*}
    \text{TRD} = \sum_{(i, j) \in E} \left| \frac{d_{ij} - d_{0,ij}}{d_{0,ij}} \right| \times 100\%,
\end{equation*}
where \( d_{0,ij} \) and \( d_{ij} \) are the initial and real-time distances between the \(i\)-th and \(j\)-th fingertips, respectively.

\begin{figure}[t]
    \centering
    \includegraphics[height=7cm]{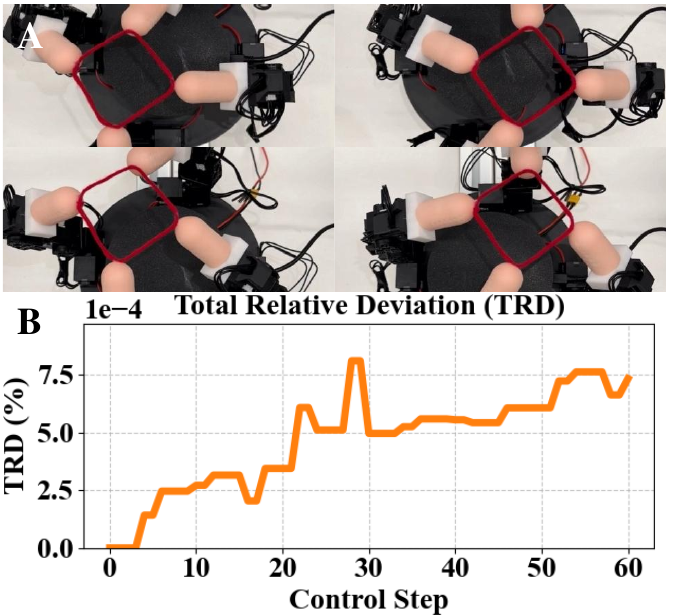}
    \caption{\small Rigid contact model validation: (A) Hardware execution showing finger motion. (B) Fingertip relative distance curves evaluating rigid contact constraints.}
    \label{fig:rigid_val}
    \vspace{-0.3cm}
\end{figure}

The trajectory is generated by an optimization that minimizes waypoint error, subject to the rigidity constraints in~\eqref{rigid constraint}. As shown in Fig.~\ref{fig:rigid_val}, the resulting inter-fingertip deviations are minimal, and the yarn remains taut without significant stretching. This result confirms that our method successfully maintains the rigid grasp throughout the task.
\subsubsection{Plastic Cup Manipulation: Validating the Force Control Strategy}
To validate our force control strategy, we use a disposable plastic cup with an AprilTag attached to its rim for visual feedback. The system guides the AprilTag along a predefined trajectory using visual and joint encoder feedback.

First, an empty plastic cup ($M=3\mathrm{g}$, compliance gains $[c_x,c_y,c_z]=[2,2,5]$, $\mu=0.65$) is guided through two waypoints: $[0,0.03,0.03]$ and $[0,-0.03,-0.02],\mathrm{m}$. Using our Algorithm~\ref{force-to-joint converter} with optimization parameters ($\lambda_1 = [30, 30, 30, 0.01, 0.01, 0.01]$, $\lambda_2 = [0.1, 0.1, 0.1]$, $\delta = 0.001$, $\epsilon = 10^{-9}$, 20 iterations, $T=2$), the cup is manipulated stably, preserving its structural integrity while smoothly tracking the trajectory.

\begin{figure}[t]
    \vspace{-0.25cm}
    \centering
    \includegraphics[height=8.9cm]{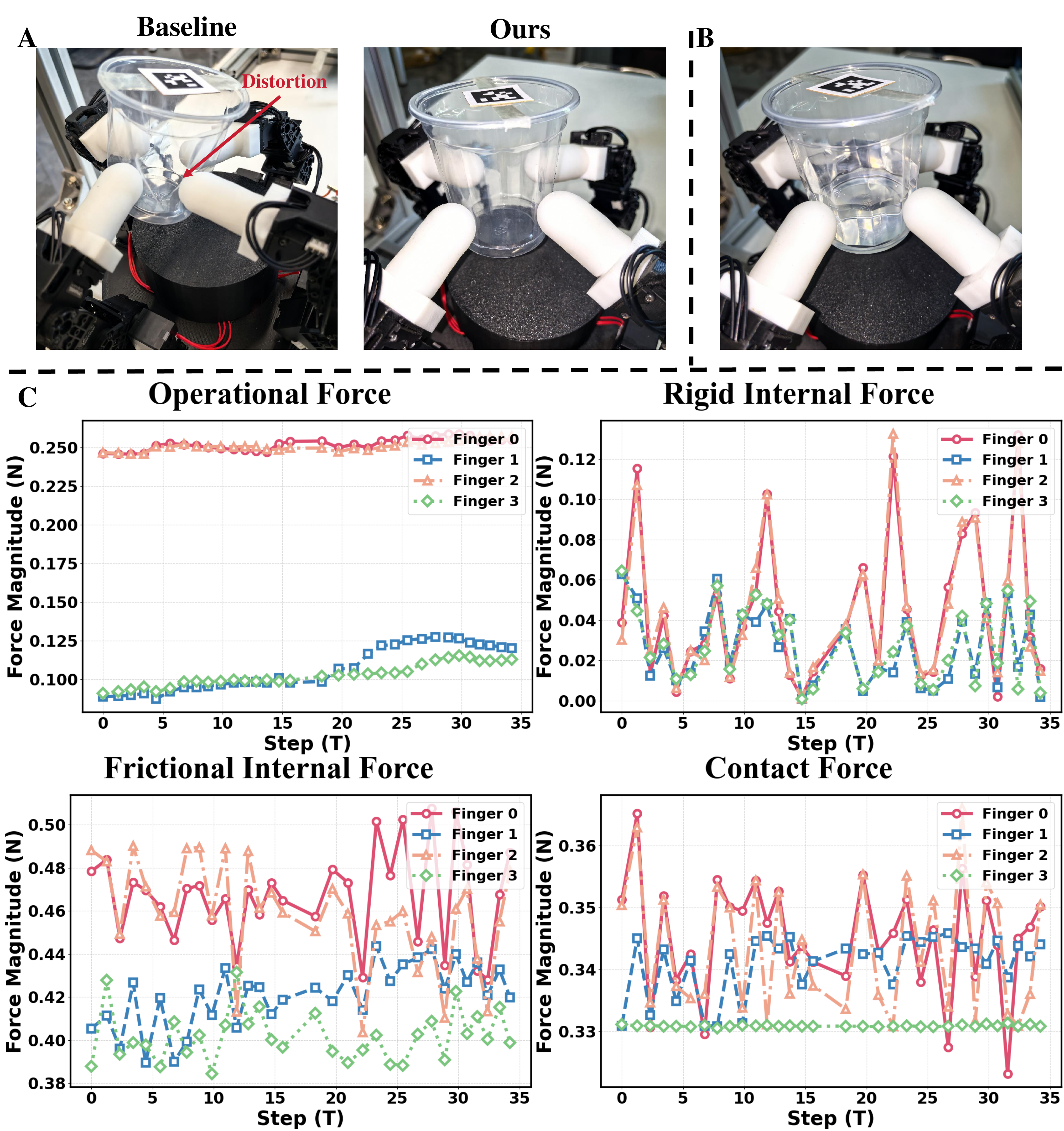}  
    \caption{\small Force control validation: (A) Empty plastic cup manipulation. The baseline deforms the cup with unstable grasp, while ours maintains stability. (B) Water-filled cup manipulation. (C) Contact forces and decompositions. The discreteness arises from MPC's step-wise nature and sparse waypoints.}
    \label{fig:force_val}
    \vspace{-0.25cm}
\end{figure}

The task difficulty is then increased using a water-filled cup (mass: $53\mathrm{g}$; compliance gains: $[3.5,3.5,9]$). A new three-waypoint trajectory is commanded: $[0,0.035,0]$, $[0,-0.035,0]$, and $[0,0,0.03],\mathrm{m}$, with other parameters held constant. As shown in Fig.~\ref{fig:force_val}, our method again achieves a stable grasp and accurate trajectory tracking, with no observable slippage or instability.
\subsubsection{Raw Egg Tracing: Validating Precision and Force Regulation}
To evaluate trajectory tracking precision, we task the manipulator with tracing the letters ``RAL'' in mid-air using a raw egg—a task demanding high fidelity. An AprilTag affixed to the egg's surface provides visual feedback. The system adjusts joint angles in real time to navigate a series of predefined waypoints that constitute each letter, thereby testing the algorithm's ability to execute complex, high-precision trajectories on a fragile object. The experimental parameters are as follows: the waypoints for writing the letters `R', `A', and `L' are set to $[0.02, -0.01, 0]$, $[0, -0.01, 0]$, $[-0.02, -0.01, 0]$, $[-0.02, 0.01, 0]$, $[0, 0.01, 0]$, $[0, -0.01, 0]$, and $[0.02, 0.01, 0]$ for `R'; $[-0.02, 0.01, 0]$, $[0, 0, 0]$, $[0.02, -0.015, 0]$, $[-0.02, 0, 0]$, $[0.02, 0.015, 0]$, $[0, 0.0075, 0]$, and $[0, -0.0075, 0]$ for `A'; and $[-0.02, 0, 0]$, $[0, 0, 0]$, $[0.02, 0, 0]$, $[0.02, 0.015, 0]$ for `L'. The egg's physical properties were set to a mass of $M=52.5\mathrm{g}$, compliance gains $[c_x, c_y, c_z]=[3, 3, 9]$, and friction $\mu=0.6$; the optimization was configured with parameters $\lambda_1 = [30, 30, 30, 0.01, 0.01, 0.01]$, $\lambda_2 = [0.1, 0.1, 0.1 ]$, $\delta = 0.001$, $\epsilon = 10^{-9}$, for 10 iterations and $T=2$.

\begin{figure}[t]
  \vspace{-0.35cm}
    \centering
    \includegraphics[height=6.9cm]{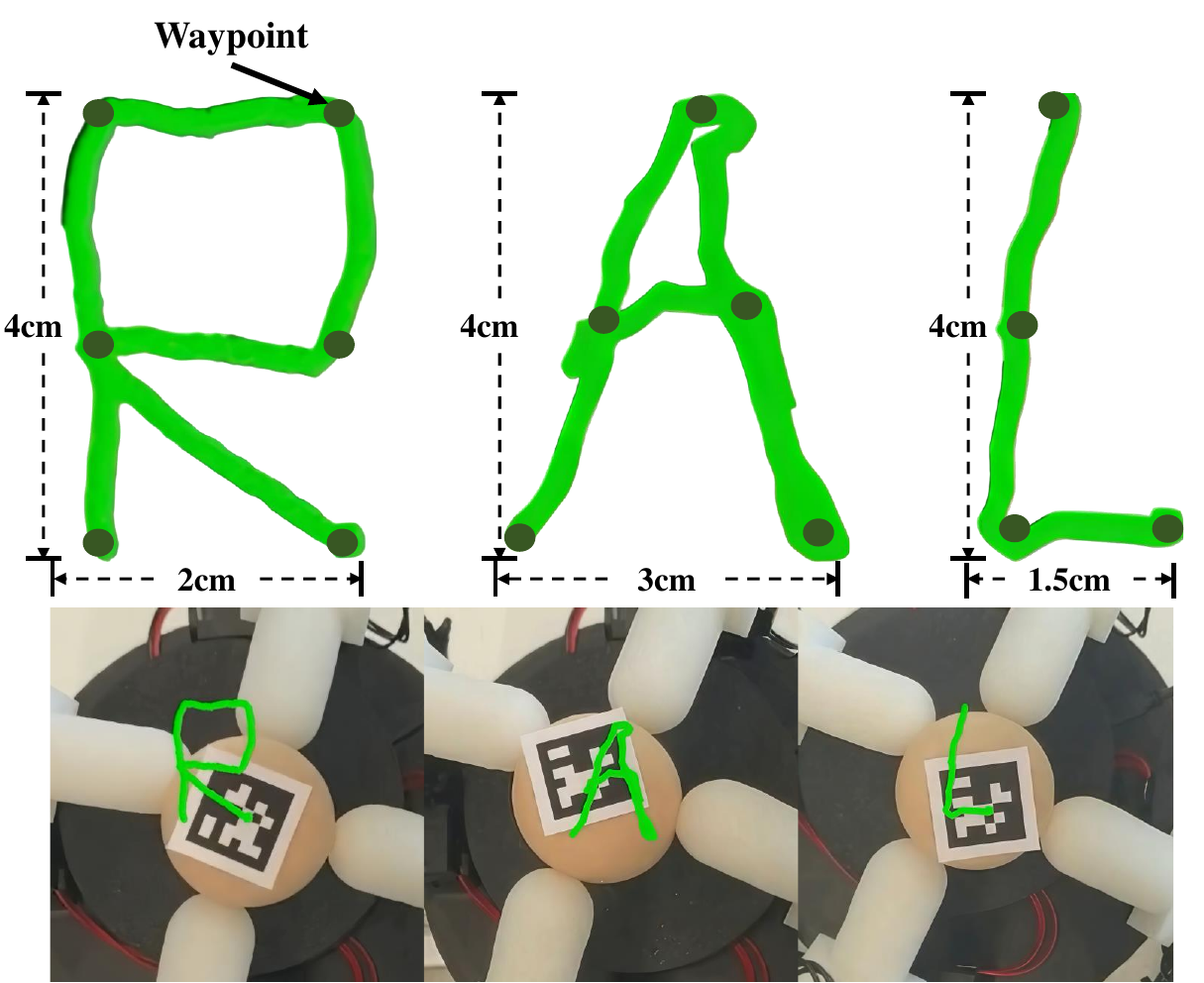}  
    \caption{\small Trajectory tracking algorithm validation experiment}
    \label{fig:tracking_val}
    \vspace{-0.3cm}
\end{figure}

Application of Algorithm~\ref{force-to-joint converter} resulted in high-precision trajectory tracking, with a maximum waypoint error of only 0.3 mm throughout the entire writing task. The execution time between consecutive waypoints ranged from 2.79 s to 6.89 s. Fig.~\ref{fig:tracking_val} visually corroborates these findings, illustrating the stable grasp and accurate path following.

\begin{figure}[ht]
  \vspace{-0.15cm}
    \centering
    \includegraphics[height=4.8cm]{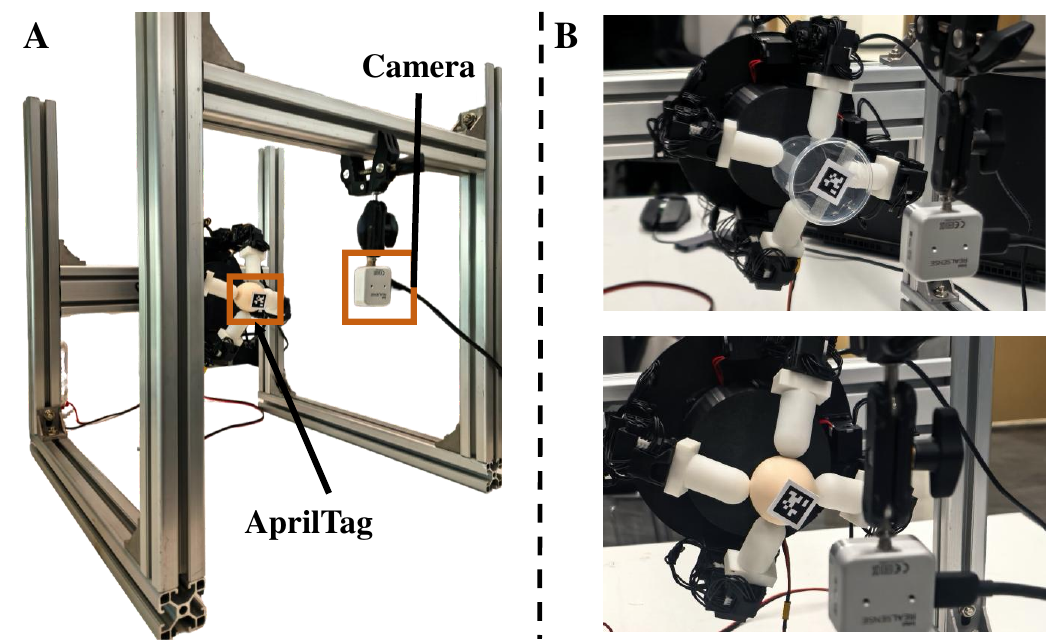}  
    \caption{\small Robustness to Gravity Direction Changes. (A) Experimental setup for vertical orientation. (B) Successful manipulation of a plastic cup (top) and a raw egg (bottom).}
    \label{fig:side_test}
    \vspace{-0.3cm}
\end{figure}

\subsubsection{Robustness to Gravity Direction Changes}
To validate robustness against gravity-induced asymmetric loading, we conducted experiments in a vertical hand orientation (side grasp). In this configuration, gravity acts tangentially to the contact surfaces, significantly increasing the risk of slippage. For the plastic cup ($[c_x, c_y, c_z] = [2, 2, 5]$, $\lambda_1 = [35, 35, 35, 0.01, 0.01, 0.01]$, $\lambda_2 = [0.1, 0.1, 0.1]$, $\delta = 0.001$, $\epsilon = 10^{-9}$, $T=2$, 25 iterations), the system tracked waypoints $[0, 0.03, 0.03]$, $[0, -0.03, -0.02]$, and $[0, 0.03, 0.03]$ with a mean error of 0.5 mm. For the raw egg ($[c_x, c_y, c_z] = [3.5, 3.5, 9.5]$, $\lambda_1 = [35, 35, 35, 0.1, 0.1, 0.1]$, $\lambda_2 = [0.1, 0.1, 0.1]$, 20 iterations), we commanded a complex trajectory: $[-0.02, 0.01, 0]$, $[0, 0.01, 0]$, $[0.02, 0.01, 0]$, $[0.02, -0.01, 0]$, $[0, -0.01, 0]$, $[0, 0.01, 0]$, and $[-0.02, -0.01, 0]$, achieving a 0.4 mm mean error. Across these tasks, execution times ranged from 3.12 s to 5.67 s per waypoint. This efficiency is competitive with the rigid object manipulation reported in \cite{hang2021manipulation}, demonstrating an effective speed-accuracy balance even under challenging gravity conditions.

\subsubsection{Robustness to Parameter Uncertainty}
\begin{figure}[ht]
  \vspace{-0.3cm}
    \centering
    \includegraphics[height=6.45cm]{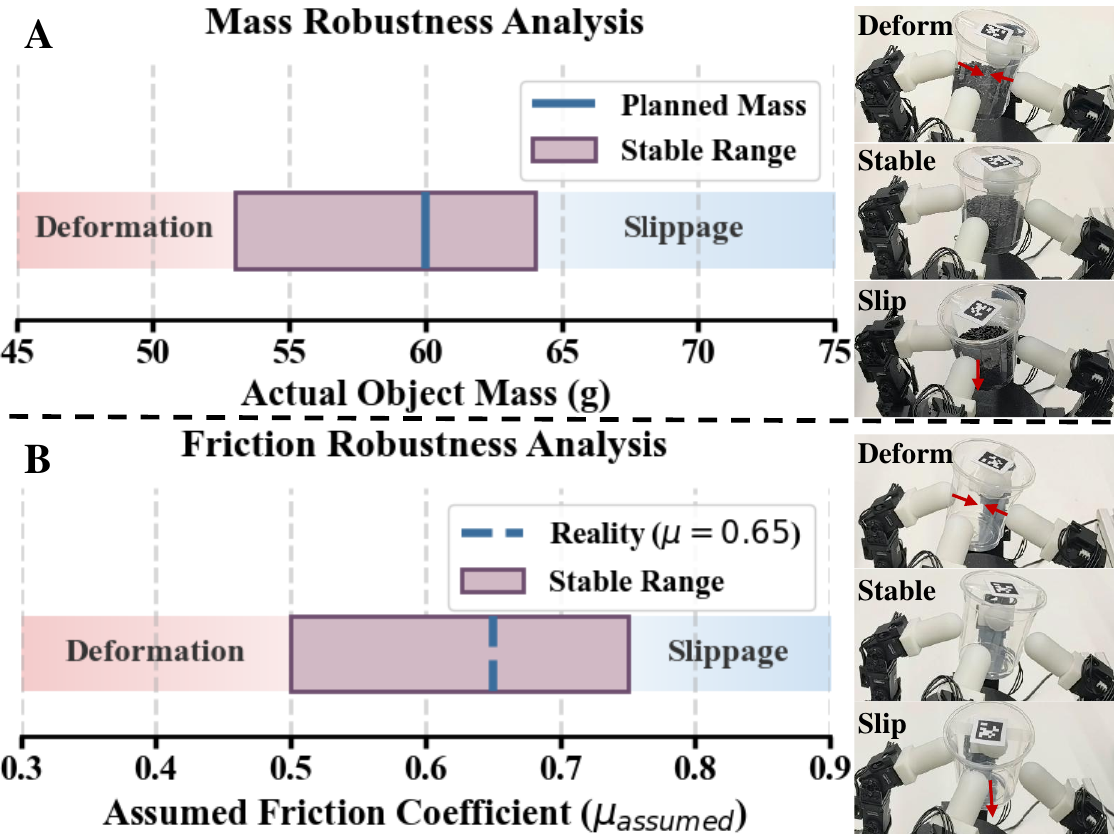}  
    \caption{\small Robustness evaluation on the disposable cup.}
    \label{fig:robustness}
    \vspace{-0.3cm}
\end{figure}
We evaluated robustness to mass ($m$) and friction ($\mu$) uncertainties by manipulating a cup filled with black rice between waypoints $p_{A,B} = [0, \pm 0.035, \pm 0.035]$.
First, regarding mass sensitivity: with the planner fixed at $m_{plan}=60$ g, the system successfully manipulated actual loads ranging from 53 g to 64 g (Fig. \ref{fig:robustness}(A)). As illustrated, lighter objects ($<53$ g) suffered deformation due to excessive internal force, while heavier ones ($>64$ g) slipped due to insufficient holding force.
Second, regarding friction estimation ($\mu_{real} \approx 0.65$): the system maintained stability even when the planner assumed $\mu_{assumed}$ values between 0.50 and 0.75 (Fig. \ref{fig:robustness}(B)).
Overall, these results confirm that the proposed framework provides a sufficient safety margin against real-world parameter uncertainties.

\section{Conclusion}
In this work, we proposed a graph rigidity-based framework for in-hand trajectory tracking with multi-finger dexterous hands, enabling precise and safe manipulation of force-sensitive objects without tactile sensing. Experiments demonstrate stable multi-finger force control.
In future work, we will use learning-based models to automatically map contact forces and penetration, and apply online system identification to estimate object parameters, reducing manual tuning and prior dependence on physical properties.

\bibliographystyle{IEEEtran}  
\bibliography{references} 
\end{document}